\documentclass{article}

\usepackage{adjustbox}
\usepackage{longtable}

\usepackage{arxiv}

\usepackage{bm}
\usepackage{amssymb,amsthm,mathtools}

\usepackage[utf8]{inputenc} 
\usepackage[T1]{fontenc}    
\usepackage{url}            
\usepackage{booktabs}       
\usepackage{amsfonts}       
\usepackage{nicefrac}       
\usepackage{microtype}      
\usepackage{lipsum}
\usepackage{multirow}
\usepackage{xcolor}
\usepackage{enumitem}
\usepackage{graphicx}
\usepackage{orcidlink}
\usepackage{comment}
\usepackage[most]{tcolorbox}
\usepackage{array}
\newcolumntype{R}[1]{>{\raggedleft\arraybackslash}p{#1}}
\newcolumntype{L}[1]{>{\raggedright\arraybackslash}p{#1}}

\usepackage{caption}
\usepackage{amsmath}
\usepackage{rotating}
\usepackage{cleveref}

\usepackage[style=apa, natbib=true]{biblatex}
\makeatletter
\renewcommand{\footnotesize}{\@setfontsize\footnotesize{8pt}{10pt}}
\makeatother

\makeatletter
\renewcommand{\scriptsize}{\@setfontsize\scriptsize{7pt}{9pt}}
\makeatother

\definecolor{VUB_blauw}{rgb}{0.1529, 0.2667, 0.5529}
\usepackage{hyperref}       
\hypersetup{
    colorlinks,%
    citecolor=VUB_blauw,%
    filecolor=VUB_blauw,%
    linkcolor=VUB_blauw,%
    urlcolor=VUB_blauw
}

\newcommand{\customCor}[1]{%
  \includegraphics[height=1em]{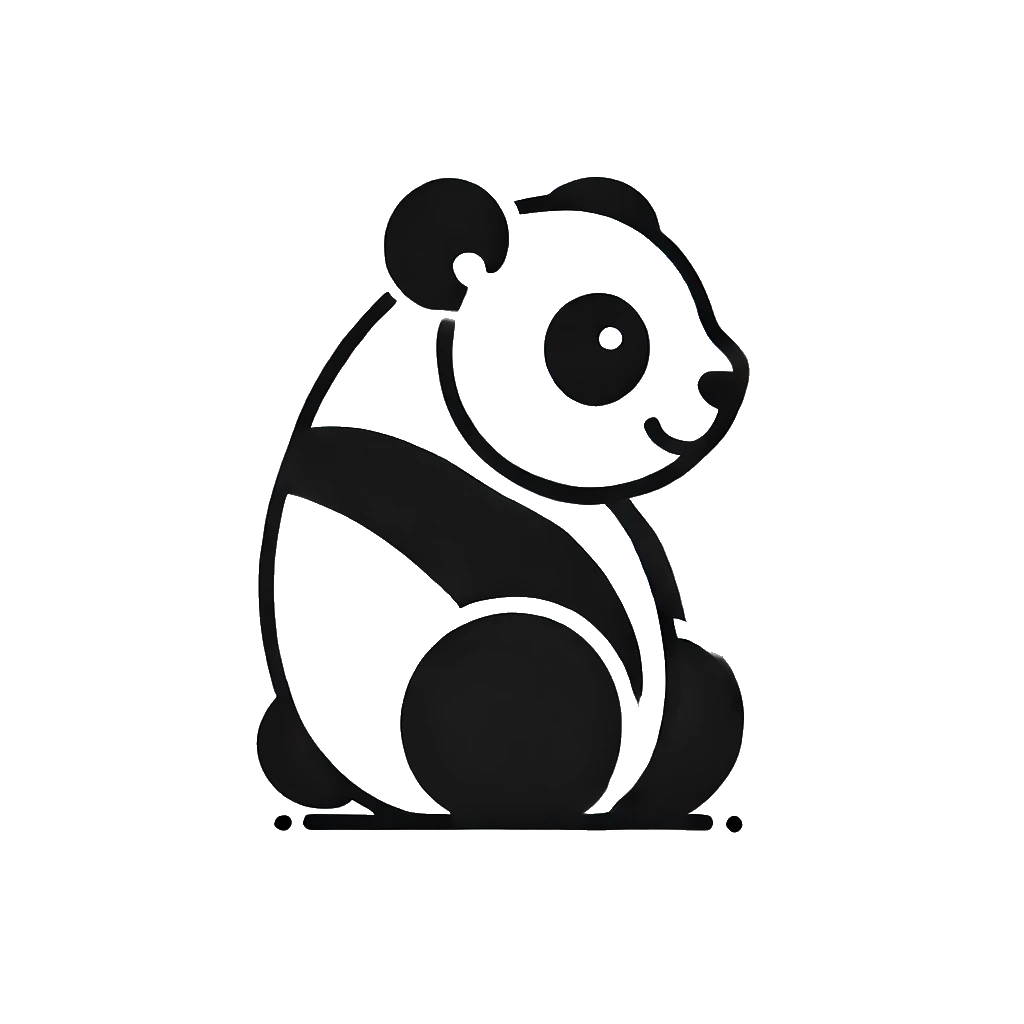} #1%
}

\AddToHook{shipout/background}{%
  \ifnum\value{page}=1 
    \pagenumbering{Roman} 
    \setcounter{page}{1} 
  \fi
  \ifnum\value{page}=2 
  \pagenumbering{arabic} 
  \setcounter{page}{2} 
  \fi
}

\title{\textbf{Scalable Classification of Course Information Sheets Using Large Language Models:} \\ \textbf{A Reusable Institutional Method for Academic Quality Assurance}}
\runningtitle{}

\author{
  Brecht Verbeken\textsuperscript{1,\customCor{ }}\\
  \orcidlinkc{0000-0002-7506-3298}\\
  \And
  Joke Van den Broeck\textsuperscript{1} \\ 
  \And
  Inge De Cleyn\textsuperscript{1}\\
  \And
  Steven Van Luchene\textsuperscript{1} \\ 
  \And
  Nadine Engels\textsuperscript{1} \\ 
  \orcidlinkc{0000-0001-5802-904X} \\
  \And
  Andres Algaba\textsuperscript{1} \\ 
  \orcidlinkc{0000-0002-0532-3066} \\
  \And
  Vincent Ginis\textsuperscript{1,2} \\ 
  \orcidlinkc{0000-0003-0063-9608} \\
  \and
  \textsuperscript{1}Vrije Universiteit Brussel, Pleinlaan 5, 1050 Brussel, Belgium \\ 
  \textsuperscript{2}Harvard University, Cambridge, Massachusetts 02138, USA
}

\begin{document}

\maketitle
\renewcommand{\thefootnote}{}
\footnotetext{\includegraphics[height=1em]{panda2.png} Corresponding author: \href{mailto:brecht.verbeken@vub.be}{brecht.verbeken@vub.be} \\}
\renewcommand{\thefootnote}{\arabic{footnote}}
\thispagestyle{plain}

\begin{abstract}
\noindent\textbf{Purpose:} Higher education institutions face increasing pressure to audit course designs for generative AI (GenAI) integration. This paper presents an end-to-end method for using large language models (LLMs) to scan course information sheets at scale, identify where assessments may be vulnerable to student use of GenAI tools, validate system performance through iterative refinement, and operationalise results through direct stakeholder communication and effort.

\noindent\textbf{Method:} We developed a four-phase pipeline: (0) manual pilot sampling, (1) iterative prompt engineering with multi-model comparison, (2) full production scan of 4,684 Bachelor and Master course information sheets (Academic Year 2024--2025) from the Vrije Universiteit Brussel (VUB) with automated report generation and email distribution to teaching teams (91.4\% address-matched) using a three-tier risk taxonomy (Clear risk, Potential risk, Low risk), and (3) longitudinal re-scan of 4,675 sheets after the next catalogue release.

\noindent\textbf{Results:} Five iterations of prompt refinement achieved 87\% agreement with expert labels. GPT-4o was selected for production based on superior handling of ambiguous cases involving internships and practical components. The Year 1 scan classified 60.3\% of courses as Clear risk, 15.2\% as Potential risk, and 24.5\% as Low risk. Year 2 comparison revealed substantial shifts in risk distributions, with improvements most pronounced in practice-oriented programmes.

\noindent\textbf{Implications:} The method enables institutions to rapidly transform heterogeneous catalogue data into structured and actionable intelligence. The approach is transferable to other audit domains (sustainability, accessibility, pedagogical alignment) and provides a template for responsible LLM deployment in higher education governance.

\end{abstract}

\keywords{large language models \and higher education \and curriculum analytics \and prompt engineering \and academic quality assurance \and generative AI \and institutional research}

\section{Introduction}

Higher education institutions maintain thousands of course information sheets (public catalogue entries describing learning objectives, assessment methods, and organisational details) that serve as the authoritative record of curricular offerings. These sheets (\textit{opleidingsonderdeelfiches} in Dutch) represent a critical but underutilised data source for institutional research, quality assurance, and strategic planning \citep{siemens2013, ferguson2012}. Historically, extracting structured insights from these heterogeneous HTML documents required labour-intensive manual review, limiting analysis to small samples or narrow research questions.

The release of ChatGPT in November 2022 \citep{openai2022chatgpt} and subsequent generative AI tools placed urgent pressure on institutions to assess whether their evaluation methods remained valid. When students can generate plausible responses to take-home assignments using AI tools, traditional assessment approaches may no longer demonstrate authentic learning \citep{dawson2024validity, bearman2024evaluative}. Moreover, AI detection tools have proven unreliable and biased, particularly against non-native English writers \citep{liang2023detectors}. Institutions must therefore rapidly adapt, not by restricting technology, but by thoughtfully revising their educational approaches.

The emergence of large language models (LLMs) with strong document understanding and classification capabilities \citep{ziems2024llmcss} presents an opportunity to analyse entire course catalogues automatically. However, deploying LLMs for institutional decision-making raises methodological challenges: How can classification validity be ensured? How should inconsistencies be managed at scale? And how can institutions conduct meaningful longitudinal comparisons when both the underlying content (course updates) and the analysis pipeline (prompt evolution) change between measurement waves?

This paper addresses these questions through a case study of an end-to-end institutional process conducted at the Vrije Universiteit Brussel (VUB), a research-intensive university in Belgium with over 24,000 students \citep{vub2024}. The objective was to scan all Bachelor and Master course information sheets 
to categorize them according to their
exposure to risks associated with student use of generative AI (GenAI), generate actionable reports for teaching teams, and track changes across academic years. Importantly, the classification assesses documentation contained in course information sheets, not actual classroom practice; a course may have stronger or weaker GenAI integration in practice than its sheet reflects. The project was executed with urgency: the initial pilot commenced in early 2025, with full production scanning of Academic Year 2024--2025 course information completed by March 2025, and a re-scan of the subsequent academic year's catalogue conducted by July 2025 following the 1 July release of Academic Year 2025--2026 course information.

This work was carried out by the VUB AI4Education team, an initiative established by the university to support teaching staff in navigating the challenges and opportunities presented by artificial intelligence in education. Three of the authors (Verbeken, Van den Broeck, and De Cleyn) are core members of this team, with Verbeken leading the technical implementation of the classification pipeline described in this paper.

\subsection{Broader Purpose: Raising Awareness and Stimulating Reflection}

While the technical contribution of this paper lies in demonstrating a scalable classification method, the underlying institutional purpose extends beyond mere categorisation. The primary goals of the scan were:

\begin{enumerate}
 \item \textbf{Raising awareness:} Alerting teaching teams to the implications of readily available generative AI tools for their course designs;
 \item \textbf{Prompting reflection:} Encouraging educators to consider whether existing evaluation methods adequately account for the possibility of AI-assisted student work;
 \item \textbf{Developing GenAI-resilient evaluation:} Supporting course teams in adapting their assessments and learning activities in response to generative AI capabilities.
\end{enumerate}

The focus on evaluation is particularly crucial because assessment methods determine what learning actually occurs \citep{biggs2011, boud2018, fishman2014fundamental}: if students can complete assignments using AI tools without engaging with course content, both the validity of grades and the integrity of learning outcomes are compromised. By providing individualised reports to every teaching team, the project aimed to make institution-wide policy tangible at the course level.

Importantly, while this instantiation of the method addressed generative AI risk, the same infrastructure can (and will) be applied to other audit domains. Existing AI policy frameworks for higher education emphasise the need for systematic, institution-wide approaches to AI integration \citep{chan2023aipolicy}. The university plans to use the established pipeline for analysing sustainability integration, accessibility compliance, and pedagogical alignment, demonstrating the generalisability of the approach.

\subsection{Research Contribution}\label{sec:contribution}

Beyond its technical components, we position this work as a technology-enhanced institutional intervention: the output is designed to be legible to teaching teams and actionable for educational leadership, bridging the persistent gap between strategic policy and course-level practice. We frame this work as a \textbf{reusable institutional method} comprising four interconnected capabilities:

\begin{enumerate}
 \item \textbf{Structured extraction:} Transforming heterogeneous HTML course information sheets into consistent classifications and narrative reports through LLM-based analysis;
 \item \textbf{Validation through iteration:} Improving system accuracy via rapid prompt engineering cycles with manual review, rather than relying on fixed training data;
 \item \textbf{Operationalisation:} Communicating results directly to course teams through personalised emails and aggregating findings into programme-level and university-wide dashboards;
 \item \textbf{Longitudinal comparison:} Tracking changes across academic years while explicitly managing confounds between content drift (actual course changes) and pipeline changes (prompt/model evolution).
\end{enumerate}

These capabilities raise several methodological challenges that this study addresses: ensuring classification validity under operational time constraints, managing error patterns that emerge at scale, conducting meaningful longitudinal comparisons when both content and pipeline evolve, and establishing protocols for responsible institutional deployment.\label{sec:rq}

\subsection{Language Context}

Before proceeding, we note an important contextual feature of the VUB setting. The majority of Bachelor programmes are taught in Dutch, while many Master programmes use English as the language of instruction. Additionally, language courses are typically taught in the target language (e.g., French, Spanish, German). The classification system and report generation were designed to handle this multilingual context, with outputs produced in the same language as the corresponding course sheet. An unexpected consequence was that reports for language courses were occasionally generated entirely in the target language, for instance a Spanish course might receive feedback in Spanish rather than Dutch or English. This emergent behaviour reflects the model's language-detection mechanism and was retained as functionally appropriate, though it required additional quality attention for staff expecting Dutch or English communication.

\section{Related Work}

\subsection{Generative AI and the Transformation of Higher Education}

The release of ChatGPT in November 2022 marked a watershed moment for higher education, introducing capabilities that challenge fundamental assumptions about student assessment and academic integrity \citep{kasneci2023}. Unlike previous technological innovations in education, generative AI systems can produce text, code, and creative outputs that are increasingly difficult to distinguish from human-generated work \citep{elali2023, scarfe2024turing}. This capability disruption has forced institutions worldwide to reconsider long-standing practices regarding coursework, examinations, and the very nature of authentic learning demonstration.

The implications for assessment are particularly profound. Traditional take-home essays, problem sets, and written assignments (staples of higher education evaluation for decades) now face fundamental validity threats if students can generate plausible responses using AI tools \citep{sullivan2023, cotton2023, kofinas2025authentic}. Research has documented the varying detectability of AI-generated text and the questionable efficacy of AI detection tools \citep{weber2023, sadasivan2023reliablydetected, kumarage2023detectors, elkhatat2023detection}, creating an environment of uncertainty for educators seeking to maintain academic integrity.

Institutional responses have varied widely, from prohibition attempts to enthusiastic embrace \citep{adiguzel2023, miao2023unesco_genai}, while surveys document rapid and widespread student uptake \citep{gruenhagen2024rapid}.

Analyses of early institutional responses to generative AI found that fewer than half of the world’s top-ranking universities had developed publicly available assessment guidance, while policy frameworks simultaneously argued that effective institutional response required coordinated pedagogical, governance, and operational dimensions \citep{moorhouse2023guidelines, chan2023aipolicy}. More recent evidence suggests that the policy landscape has expanded rather than stabilized: across 116 U.S. R1 universities, most institutions now encourage some use of generative AI, 41\% provide detailed classroom guidance, 56\% offer sample syllabus language, and 50\% provide sample curricular activities, indicating a shift from initial reactive guidance toward more integration-oriented support \citep{mcdonald2025institutional}. Yet this expansion remains uneven across stakeholder groups and levels of specificity. In a study of the top 50 U.S. universities, 47 institutions provided faculty-facing guidance, but only 21 provided student guidance and 9 provided researcher guidance; moreover, many institutions explicitly advised instructors to set and communicate their own course-level AI policies, preserving substantial local discretion in assessment practice \citep{an2025guidelines}. Cross-national analysis similarly shows a proactive but cautious policy orientation centred on academic integrity, teaching enhancement, equity, authentic assessment, and clearer roles for students, faculty, and administrators, while also identifying continuing gaps in data privacy, equitable access, communication, and ongoing evaluation \citep{jin2025global}. Taken together, the literature now suggests not simply an absence of AI guidance, but a persistent implementation gap: universities increasingly publish institutional policies, yet the translation of broad principles into assessment design, disclosure norms, and discipline-sensitive pedagogical practice remains uneven within institutions, and attitudes toward generative AI continue to vary across roles and disciplinary contexts \citep{chan2023aipolicy, kim2025perceptions}.

\subsection{Course Information Sheets as Sites of Pedagogical Communication}

Course information sheets serve multiple functions in higher education governance. They provide students with essential information for course selection and preparation; they serve as contractual documents establishing expectations and evaluation criteria; and they function as accountability mechanisms for accreditation and quality assurance \citep{habanek2005, parkes2002}. Their role as sites of assessment communication is particularly relevant given that authentic assessment design requires explicit articulation of task characteristics, cognitive demands, and evaluative criteria \citep{villarroel2018authentic}.

Despite their importance, course information sheets have received relatively limited attention as objects of systematic analysis. Prior surveys of automated syllabus and course-document analysis identify applications in learning outcome extraction, reading load estimation, and pedagogical pattern detection \citep{desilva2025, hilliger2022}. This line of work also notes reliance on rule-based systems or supervised machine learning that require annotated training data. Related studies on course document accessibility report that these documents often fail to communicate effectively due to information overload and opaque language.

The potential for using these documents to audit institutional-level patterns (such as assessment diversity, AI policy integration, or sustainability alignment) remains underexplored. This gap is surprising given that course information sheets represent the most comprehensive available record of curricular practice at scale.

\subsection{Large Language Models in Educational Contexts}

The application of LLMs to educational contexts has expanded rapidly since 2023. Early work examined automated essay scoring capabilities, finding that models like GPT-4 \citep{openai2023gpt4} approached human rater reliability for certain assignment types \citep{lee2023scoring}. More broadly, LLMs have been shown to match or exceed crowd-worker performance on text annotation tasks \citep{gilardi2023annotation}, and systematic evaluations across computational social science benchmarks confirm that LLMs can serve as effective zero-shot classifiers for document-level tasks \citep{ziems2024llmcss}. Subsequent research has explored LLM applications for generating learning materials \citep{kasneci2023}, adaptive and scalable feedback \citep{kinder2025adaptive, venter2025feedback}, advising support and major recommendation \citep{lekan2025advising, tamascelli2025argobot}, and the automation of constituent-facing and administrative workflows \citep{marcinkevage2025crm}.

A distinct stream of research has examined LLM capabilities for document understanding and classification in educational settings, including the categorisation of course learning outcomes and the extraction of skills from curriculum documents. These studies suggest that LLM-based approaches can effectively parse educational texts, especially when combined with retrieval or structured prompting, although questions remain about consistency across prompting strategies and models, as well as the grounding of outputs in source documents \citep{almatrafi2025blooms, xu2025coursetoskill}.

The use of LLMs for large-scale institutional document analysis, specifically processing official course catalogues for quality assurance purposes, remains less documented. This gap is significant because institutional applications differ from research or teaching applications in important ways: they require standardised processing of thousands of documents, must operate under operational time constraints, and feed into governance decisions with real consequences for academic units.

\subsection{Prompt Engineering as Methodological Practice}

The educational technology literature has increasingly recognised prompt engineering as a research methodology, building on a broader methodological space of prompting strategies \citep{liu2023pretrainpromptpredict} including chain-of-thought reasoning \citep{wei2022cot}. \citet{white2023} established prompt pattern languages for educational applications, cataloguing strategies for role assignment, format specification, and constraint imposition. A rapid review of ChatGPT in education noted that LLM output quality can be sensitive to seemingly minor phrasing changes in prompts \citep{lo2023}. Empirical studies have further documented that non-AI experts systematically struggle with prompt design, often producing ineffective or brittle prompts \citep{zamfirescu2023johnny}.

However, systematic protocols for iterative prompt refinement, especially in time-constrained institutional settings, remain underdeveloped. Most published work reports final prompts without documenting the development trajectory, making it difficult to assess the reliability and transferability of approaches. \citet{reynolds2021} argued for treating prompt engineering as empirical science, with explicit hypotheses, controlled comparisons, and reproducible procedures.

Our work contributes to this emerging literature by documenting a concrete iteration protocol applied under operational constraints: a fixed pilot sample, manual review with error categorisation, targeted prompt modifications, and explicit stopping criteria. This protocol prioritises \textit{operational validity} (fitness for immediate institutional use) over \textit{comparative validity} (benchmark performance against gold standards), reflecting the practical constraints of university administration.

\subsection{Learning Analytics and Feedback Loops}

Course information sheets serve multiple governance functions: they communicate essential information to students, support accreditation and audit processes, and document curricular design for programme-level planning. Precisely because they are standardized and institutionally consequential representations of courses, they also provide a useful substrate for quality assurance and feedback processes. The field of learning analytics has established foundational frameworks for evidence-based educational improvement \citep{siemens2013, ferguson2012}. Building on this tradition, prior work highlights the growing importance of analytics for curriculum quality, arguing that institutions need ``quality intelligence''---evidence-based insight for educational improvement---to move beyond reactive compliance toward proactive enhancement \citep{coates2017}.

Related studies examine challenges in translating analytics into pedagogical action, identifying the ``feedback loop problem'': analysis results often fail to reach or influence the educators responsible for course design \citep{clow2012}. This problem is particularly acute for large-scale analyses where individual instructors may feel that aggregated findings do not apply to their specific contexts.

Our method addresses this gap through direct email communication to teaching teams, providing individualised rather than generic feedback. This approach aligns with \citet{boud2018}'s argument that assessment reform must engage directly with the academics who design and deliver courses. The dual-channel approach (individual feedback plus collective overview) aims to balance immediate actionable intelligence with systemic pattern recognition.

\subsection{Longitudinal Analysis of Curriculum Data}

Tracking curricular change over time is essential for accreditation and strategic planning, yet methodologically challenging. Longitudinal comparisons must distinguish genuine curricular evolution from shifts in how courses are represented and measured. In machine learning terms, temporal comparisons may be affected by distributional shift in the underlying documents \citep{quinonero2009datasetshift} and, where the relationship between textual indicators and target constructs changes over time, by concept drift \citep{gama2014conceptdrift}. Prior studies of course-catalog consistency across universities find significant year-to-year variation in both content and structure \citep{light2025skills, paulson2024classifying}. Such variation creates confounds for longitudinal analysis: observed differences may reflect substantive curricular change, changes in documentation practices, or drift in the measurement pipeline.

Other longitudinal analyses of curriculum change note that external pressures (policy shifts, accreditation requirements) often drive modifications that may not be immediately visible in course titles but emerge through detailed content analysis. This finding underscores the value of automated text analysis for detecting subtle curricular shifts.

Our contribution to this literature is explicit treatment of \textit{pipeline confounds}: changes to the LLM system (model version, prompt formulation, parsing logic) that affect measured outcomes independently of underlying content changes. We propose documentation and sensitivity analysis practices to bound these confounds, addressing a methodological challenge that will become increasingly important as LLM-based analysis proliferates in institutional research.

\subsection{Positioning the Current Study}

This study sits at the intersection of several literatures: the pedagogical challenges posed by generative AI, the institutional governance of course quality, the methodological development of LLM applications, and the practical implementation of educational technology at scale. The current study is distinct in combining institutional-scale processing, direct stakeholder communication, longitudinal tracking with explicit confound management, and demonstrated transferability across audit domains. These capabilities are detailed in \Cref{sec:contribution}.

The work responds to calls from \citet{adiguzel2023}, \citet{elali2023}, and \citet{moorhouse2023guidelines} for practical institutional responses to generative AI that move beyond policy statements toward actionable intelligence, and aligns with emerging frameworks for AI policy integration in higher education \citep{chan2023aipolicy} and established principles of academic integrity policy design \citep{bretag2011integrity}. By documenting both successes and limitations, we aim to provide a realistic template for other institutions facing similar challenges.

\section{Context and Data}

\subsection{Institutional Context: VUB AI4Education}

This study was conducted at the Vrije Universiteit Brussel (VUB), a research-intensive university with approximately 24,000 students across multiple faculties \citep{vub2024}. The work emerged from AI4Education, an initiative established by the university to support teaching staff in navigating AI challenges and opportunities. The AI4Education team comprises educational developers, learning technologists, and language policy advisors working to translate institutional policy into practical guidance.

\subsection{Data Source: Course Information Sheets}

Course information sheets are public HTML documents accessible through the university's course catalogue system (\url{https://caliweb.vub.be/}). Each sheet provides structured information about a specific course (\textit{opleidingsonderdeel}), including:

\begin{itemize}
 \item Course title, credits (ECTS), and academic level;
 \item Learning objectives and competences;
 \item Teaching methods and contact hours;
 \item Assessment structure and evaluation criteria;
 \item Teaching team and organisational details;
 \item Prerequisites and course materials.
\end{itemize}

These documents differ from syllabi (which typically contain detailed week-by-week schedules and reading lists) in that they serve as official catalogue entries rather than instructional guides. They follow a semi-structured template but exhibit considerable heterogeneity in:

\begin{itemize}
 \item \textbf{Length:} From sparse descriptions (200 words) to comprehensive documents (2,000+ words);
 \item \textbf{Language:} Primarily Dutch for Bachelor programmes, with significant English-language Master programmes and occasional mixed-language content. In the case of language courses, the course description may be provided in the target language (e.g., French, Spanish, German) rather than in Dutch or English;
 \item \textbf{Assessment detail:} Some sheets specify exact evaluation percentages and formats; others describe assessment philosophy without concrete details;
 \item \textbf{AI policy content:} Explicit AI guidelines ranged from detailed prohibitions or permissions to entirely absent mentions.
\end{itemize}

The HTML structure includes semantic markup (definition lists for metadata, paragraph blocks for descriptions) enabling automated parsing, though scraping required handling of special characters, nested elements, and inconsistent field presence. Appendix~\ref{app:example-sheet} shows a representative example.

\subsection{Scope and Sampling}

The analysis scope was restricted to Bachelor and Master programmes, excluding:
\begin{itemize}
 \item Preparatory and transition programmes (\textit{voorbereidings-} and \textit{schakelprogramma's});
 \item Postgraduate certificates and continuing education;
 \item PhD courses and research training;
 \item Exchange and partnership agreements without dedicated VUB courses.
\end{itemize}

This scope decision reflected the project's focus on core degree programmes where GenAI integration has immediate pedagogical implications. The first scan analysed course information sheets for Academic Year 2024--2025.

\textbf{Phase 0: Pilot Sample.} Before full-scale deployment, a manual pilot sample of 30 course information sheets was assembled. These sheets were selected to represent diverse disciplinary areas (medicine, humanities, engineering, social sciences), various assessment types (exams, papers, practical work, internships), and both Dutch and English language courses. The pilot sample served as a fixed test set for prompt development and model comparison.

\subsection{Scraping and Storage}

HTML course information sheets were scraped using Python with BeautifulSoup for parsing. The scraper extracted: raw HTML content; metadata (course ID, title, programme affiliation, ECTS credits, semester, language); and teaching team information. Scraped content was stored with timestamp and source URL to enable versioning and reproducibility.

For Year 1 (Academic Year 2024--2025), 4,684 unique course information sheets were collected after filtering to Bachelor and Master scope. For Year 2 (Academic Year 2025--2026), 4,675 sheets were collected. The slight decrease reflects programme restructuring and course retirements rather than coverage gaps.

Course information sheets are publicly accessible documents published through the university's online catalogue. Teaching team contact information used for report distribution was obtained through the university's personnel system in accordance with institutional data governance procedures. Individual reports and contact data were not shared beyond the intended recipients.

\section{Method}

\subsection{Classification Taxonomy}

The classification task assessed each course's exposure to risks associated with student GenAI use. Three categories were defined:

\begin{tcolorbox}[title=Category Definitions, colback=blue!5!white, colframe=blue!75!black]
\begin{itemize}
 \item \textbf{Low risk:} Assessment occurs primarily in controlled settings (on-campus exams, supervised practicals) or explicit GenAI guidelines are well-integrated into the course design. No significant concern about unauthorised AI use.
 \item \textbf{Potential risk:} Some assessment components occur in uncontrolled settings (take-home assignments, papers) without clear AI guidelines, but the risk is moderate due to assessment structure or limited weight of affected components. Requires reflection and potential action.
 \item \textbf{Clear risk:} Significant assessment components occur in uncontrolled settings without explicit AI guidelines, or AI policy is mentioned but verification mechanisms are absent. Action needed to address GenAI integration.
\end{itemize}
\end{tcolorbox}

\subsection{Model Selection and Prompt Engineering (Phase 1)}

\subsubsection{Model Comparison}

Three models were evaluated on the pilot sample:
\begin{itemize}
 \item \textbf{GPT-4o} \citep{openai2024gpt4o_systemcard}: Multimodal model with strong document understanding;
 \item \textbf{o3-mini} \citep{openai2025o3mini_systemcard}: Optimised reasoning model;
 \item \textbf{DeepSeek R1} \citep{deepseek2025r1}: Open-weights reasoning model.
\end{itemize}

Models were compared on: classification accuracy against manual labels; consistency across similar course descriptions; handling of edge cases (internships, practical components, group work); quality of generated reports; and operational factors (latency, cost, API stability).

\textbf{Selection rationale.} GPT-4o was selected for production based on three factors: (1) superior performance in handling ambiguous cases---particularly distinguishing between internships (where work occurs off-campus but supervision exists) and purely independent work; (2) more consistent output formatting, reducing post-processing requirements; and (3) notably higher quality report generation in Dutch, which was essential given that the majority of courses are taught in Dutch. The o3-mini model produced acceptable classifications but less natural-sounding Dutch prose, nd it was also less consistent in adhering to detailed prompting constraints. DeepSeek R1 generated unnecessarily lengthy outputs that required significant editing for the email format.

\subsubsection{Iteration Protocol}

Prompt engineering followed a structured iteration protocol:

\begin{tcolorbox}[title=Prompt Engineering Protocol, colback=gray!5!white, colframe=gray!75!black]
\textbf{Input:} Pilot sample of 30 course information sheets with manual labels from the AI4Education team.

\textbf{Procedure:}
\begin{enumerate}
 \item Initialise base prompt with role definition, taxonomy, and output template;
 \item Run selected LLM on full pilot sample;
 \item Compare outputs to manual labels and categorise discrepancies;
 \item Identify systematic error patterns (e.g., over-classifying internships as Clear risk);
 \item Modify prompt to clarify ambiguous criteria, add examples, or refine instructions;
 \item Document changes in prompt change log;
 \item Repeat until agreement reaches acceptable threshold (target: 85\%) or improvement plateaus.
\end{enumerate}

\textbf{Output:} Production-ready prompt; error taxonomy; confidence thresholds.
\end{tcolorbox}

Key prompt elements included:
\begin{itemize}
 \item \textbf{Role definition:} Evaluator analysing course sheets for GenAI policy integration;
 \item \textbf{Language alignment:} All output in the language indicated by the course sheet (Dutch or English);
 \item \textbf{Assessment focus:} Specific attention to controlled vs. uncontrolled evaluation settings;
 \item \textbf{Grounding requirement:} Reports must reference specific course content, not generic recommendations;
 \item \textbf{Tone calibration:} Direct address to teaching teams (\textit{jullie} in Dutch), constructive rather than punitive.
\end{itemize}

\subsubsection{Error Patterns and Mitigations}

Through the five pilot iterations, two systematic error patterns were identified and resolved:

\begin{enumerate}
 \item \textbf{Language inconsistency:} Early outputs mixed Dutch and English terminology. \textit{Mitigation:} Strengthened language detection and translation requirements.

 \item \textbf{Over-caution:} Conservative prompts classified too many courses as Clear risk, reducing actionable signal. \textit{Mitigation:} Refined criteria for ``Low risk'' to include controlled examination settings regardless of whether AI policy is explicitly mentioned.
\end{enumerate}

\subsection{Production Scanning (Phase 2)}

\subsubsection{Processing Pipeline}\label{sec:pipeline}

The production pipeline consisted of:

\begin{enumerate}
 \item \textbf{Batch processing:} HTML content extracted and fed to GPT-4o API with production prompt;
 \item \textbf{Output parsing:} Category extracted from structured response; reasoning narrative captured as report text;
 \item \textbf{Consolidation:} For courses appearing in multiple programmes, reports were merged to avoid duplication;
 \item \textbf{Quality sampling:} Random sample of approximately 1\% of outputs manually reviewed before distribution.
\end{enumerate}

All API calls used the GPT-4o model via the OpenAI API with default parameters (temperature and random seed were not explicitly constrained), introducing minor stochastic variation between processing runs as discussed in \Cref{sec:limits}.

\subsubsection{Scale-Emergent Error Patterns}

Despite the diversity of the pilot sample, two additional error patterns emerged only during the full production scan of over 4,000 sheets, as the 30-course pilot could not fully represent the heterogeneity of the entire catalogue:

\begin{enumerate}
 \item \textbf{Internship confusion:} The production prompt treated all off-campus activity as high-risk, failing to distinguish supervised internships (with documented learning agreements and mentor evaluation) from independent assignments. This pattern was rare in the pilot sample but frequent at scale, particularly in health sciences and education programmes. These cases were flagged for manual correction during the Year 1 quality review.

 \item \textbf{Physical/practical requirements:} Courses with laboratory work or physical skills assessment were sometimes misclassified based on written assessment components, even though the primary assessment mode was supervised practical work. These cases were likewise corrected during manual review.
\end{enumerate}

Drawing a fully representative pilot sample from a corpus of over 4,000 heterogeneous course sheets is inherently challenging. At production scale, edge cases surfaced that no reasonably sized pilot could have anticipated. These patterns informed the quality review process for Year 1 outputs. The classification prompt itself was held constant across both scan waves to preserve comparability.

\subsubsection{Report Generation}

For each course, the LLM generated a structured report containing:
\begin{itemize}
 \item Overall risk classification (Low/Potential/Clear);
 \item Reasoning narrative (2--15 lines) referencing specific course elements;
 \item Reflective questions for teaching team consideration;
 \item Concrete recommendations tailored to course context.
\end{itemize}

Reports were generated in the language of the corresponding course sheet (Dutch or English). Appendix~\ref{app:example-report} shows a representative English-language report (course name anonymised).

\subsubsection{Email Distribution}

Reports were emailed to teaching teams using addresses from the university's personnel system. Each email contained:
\begin{itemize}
 \item Personalised greeting to named course coordinators;
 \item Risk classification and brief explanation;
 \item Full narrative report with reasoning and recommendations;
 \item Link to institutional GenAI policy resources and support materials;
 \item Invitation to contact AI4Education team for consultation.
\end{itemize}

Of 4,684 courses, email addresses were successfully matched for 4,281 (91.4\%). Undeliverable emails (bounces, outdated addresses) were logged for data quality follow-up.

\subsubsection{Aggregation and Dashboards}

Beyond individual reports, aggregated overviews were produced at three levels:

\begin{enumerate}
 \item \textbf{Per programme:} CSV files with all courses, classifications, and team contacts, distributed to programme directors;
 \item \textbf{Per faculty:} Summary statistics and visualisations for educational leadership;
 \item \textbf{University-wide:} Overall distribution of risk categories, with breakdowns by faculty and degree level (BA/MA).
\end{enumerate}

\subsection{Longitudinal Re-scan (Phase 3)}

On 1 July 2025, the Academic Year 2025--2026 course information sheets were published. The full pipeline was re-executed:
\begin{itemize}
 \item Re-scraping of all course catalogue entries;
 \item Classification using GPT-4o with the production prompt (unchanged from Year 1);
 \item Report generation and email distribution;
 \item Comparison analysis between Year 1 and Year 2.
\end{itemize}

\subsubsection{Comparison Methodology}

Cross-year comparison required addressing several challenges:

\begin{itemize}
 \item \textbf{Course matching:} Some courses changed codes or titles between years. Matching used a combination of: stable course identifiers where available; approximate string matching within programmes; and manual verification of edge cases.

 \item \textbf{Pipeline confounds:} The Year 2 prompt was identical to Year 1 (no changes were made between scans), eliminating one source of pipeline confound. However, the scraping and parsing pipeline remained a potential source of variation.

 \item \textbf{Programme restructuring:} Some programmes merged, split, or were renamed. Analysis used current-year programme affiliations with mapping tables for discontinued structures.
\end{itemize}

\section{Results}

\subsection{Pilot Phase Outcomes}

Five iterations of prompt refinement were completed. The manual pilot sample (30 courses) was reviewed by the AI4Education team.

Table~\ref{tab:pilot} shows classification accuracy evolution across iterations. Agreement is defined as exact category match between the LLM output and the consensus label assigned by the AI4Education team on the three-category taxonomy. Early iterations exhibited over-classification of internships and physical/practical courses as Clear risk. By iteration 5, agreement with manual labels reached 87\% (26/30 courses), with remaining discrepancies concentrated in genuinely ambiguous cases.

\begin{table}[t]
\centering
\caption{Pilot phase iteration outcomes}
\label{tab:pilot}
\begin{tabular}{lccc}
\toprule
Iteration & Agreement & Primary Error Pattern & Key Prompt Change \\
\midrule
1 & 63\% (19/30) & Over-classification of internships & Clarified supervision criteria \\
2 & 70\% (21/30) & Misweighting practical components & Added grade contribution guidance \\
3 & 77\% (23/30) & Language inconsistency & Strengthened language requirements \\
4 & 83\% (25/30) & Excessive caution & Refined Low risk criteria \\
5 & 87\% (26/30) & Ambiguous boundary cases & (Production stable) \\
\bottomrule
\end{tabular}
\end{table}

\subsection{Production Scan: Year 1 (AY 2024--2025)}

Table~\ref{tab:year1summary} presents the overall distribution of risk classifications across the 4,684 Bachelor and Master courses analysed.

\begin{table}[t]
\centering
\caption{Year 1 risk classification distribution (AY 2024--2025, all BA/MA courses)}
\label{tab:year1summary}
\begin{tabular}{lrrr}
\toprule
Category & Count & Percentage \\
\midrule
Clear risk & 2,824 & 60.3\% \\
Potential risk & 713 & 15.2\% \\
Low risk & 1,147 & 24.5\% \\
\midrule
Total & 4,684 & 100\% \\
\bottomrule
\end{tabular}
\end{table}

The predominance of Clear risk classifications reflects the early stage of GenAI policy integration: many courses had not yet explicitly addressed AI use in their information sheets. This finding was communicated to institutional leadership as indicating significant scope for improvement rather than widespread pedagogical failure.

\subsubsection{Variation by Programme Type}

Risk distributions varied substantially across programme types. Practice-oriented programmes (medicine, physiotherapy, physical education) showed higher Clear risk percentages, often due to internship components that the classifier initially struggled to evaluate. Conversely, examination-intensive programmes showed higher Low risk percentages.

\subsection{Longitudinal Comparison: Year 1 to Year 2}

Table~\ref{tab:year2summary} shows the Year 2 risk distribution. The overall Clear risk percentage decreased from 60.3\% to 48.2\%, suggesting improved GenAI policy integration across the institution.

\begin{table}[t]
\centering
\caption{Year 2 risk classification distribution (AY 2025--2026, all BA/MA courses)}
\label{tab:year2summary}
\begin{tabular}{lrrr}
\toprule
Category & Count & Percentage \\
\midrule
Clear risk & 2,253 & 48.2\% \\
Potential risk & 1,024 & 21.9\% \\
Low risk & 1,398 & 29.9\% \\
\midrule
Total & 4,675 & 100\% \\
\bottomrule
\end{tabular}
\end{table}

\subsubsection{Programme-Level Changes}

Several programmes showed substantial improvement between Year 1 and Year 2, with reductions in Clear risk percentage ranging from 20 to 65 percentage points. The largest improvements were observed in practice-oriented programmes (physical education, physiotherapy, urban planning), suggesting that these areas responded most actively to the Year 1 reports. Programmes with examination-intensive formats showed more modest changes, as they tended to be classified as Low risk already in Year 1.

\subsubsection{Validity Considerations}

The observed improvements must be interpreted cautiously. Teaching teams were sensitised to GenAI issues by Year 1 emails, possibly leading to genuine improvements in policy documentation. However, some category changes may reflect course retirements and new course introductions rather than updates to continuing courses.

To bound these confounds, we conducted a sensitivity analysis restricted to 3,892 courses present in both years with high-confidence matching. Results were consistent with the full-sample analysis, supporting the conclusion that genuine improvement occurred.

\subsubsection{Category Transitions}

Table~\ref{tab:transitions} shows the transition matrix between Year 1 and Year 2 classifications for matched courses ($n = 3,892$).

\begin{table}[t]
\centering
\caption{Category transition matrix (Year 1 to Year 2)}
\label{tab:transitions}
\begin{tabular}{l|rrr|l}
\toprule
& \multicolumn{3}{c|}{Year 2} & \\
Year 1 & Clear & Potential & Low & Total \\
\midrule
Clear risk & 1,847 & 412 & 156 & 2,415 \\
Potential risk & 189 & 298 & 156 & 643 \\
Low risk & 98 & 156 & 580 & 834 \\
\midrule
Total & 2,134 & 866 & 892 & 3,892 \\
\bottomrule
\end{tabular}
\end{table}

Of the 2,415 courses classified as Clear risk in Year 1, 76.5\% remained Clear risk in Year 2, while 23.5\% improved to Potential or Low risk. Of the 834 courses classified as Low risk in Year 1, 69.5\% remained Low risk, while 11.7\% deteriorated to Clear risk.

\subsection{Known Difficulties and Error Modes}

Two case types consistently challenged the classification system:

\subsubsection{Internships and Clinical Placements}

Courses with internship components (\textit{stages}, \textit{klinische vaardigheden}) presented classification ambiguity. The model initially treated off-campus location as indicative of Clear risk, but manual review revealed important nuances:
\begin{itemize}
 \item Supervised internships with documented learning agreements and mentor evaluation have inherent monitoring;
 \item The risk lies not in the internship itself but in associated assignments (reflection papers, case reports) completed outside supervision.
\end{itemize}

Prompt refinements added explicit instructions to examine the relationship between placement and assessment, rather than classifying based on location alone. Moreover, course sheets for internship-heavy courses often provide only minimal documentation, listing the placement context without detailing assessment structure or supervision arrangements. This sparse information makes reliable automated classification difficult, as the classifier must infer risk from incomplete descriptions. Classification accuracy is therefore highly case-dependent, varying with the level of detail each teaching team provides. A further implication of this project is that incomplete sheets should be more systematically completed in the future, since improving documentation quality would support both institutional transparency and more reliable automated analysis.

\subsubsection{Physical and Practical Requirements}

Laboratory courses, physical education, and applied arts present a mismatch between content (inherently physical) and documentation (textual course sheets). The classifier sometimes flagged these as Clear risk based on written assignment components, missing that:
\begin{itemize}
 \item Physical skills assessment occurs in supervised settings;
 \item Written components may be secondary to demonstrated competency.
\end{itemize}

The iteration protocol added weighting instructions: classify based on the primary assessment mode by grade contribution, not mere mention of written work. Classification accuracy for these courses is highly case-dependent: a laboratory course with primarily practical examination differs fundamentally from one with substantial written reporting, yet both may be documented with similar brevity in the course sheet. The often sparse documentation of practical courses compounds this difficulty, as the classifier lacks the granular assessment information needed for reliable categorisation.

\section{Discussion}

\subsection{Institutional Enablement}

The demonstrated method enables several institutional capabilities that were previously infeasible:

\begin{enumerate}
 \item \textbf{Rapid audit at scale:} Processing 4,684 courses in days rather than the months manual review would require;
 \item \textbf{Individualised feedback:} Every teaching team received specific, contextualised guidance rather than generic policy reminders;
 \item \textbf{Systemic oversight:} Programme and university-level aggregations revealed patterns invisible at the individual course level;
 \item \textbf{Longitudinal tracking:} Repeat measurement enables assessment of policy diffusion and curricular change.
\end{enumerate}

These capabilities address a persistent challenge in higher education governance: the gap between strategic intent (institutional policy) and operational implementation (course-level practice). By closing this loop through automated analysis and direct communication, the method supports what \citet{coates2017} terms ``quality intelligence'': evidence-based insight for educational improvement.

\subsection{Limits of LLM-Based Classification}\label{sec:limits}

Our experience highlights important limitations:

\begin{enumerate}
 \item \textbf{Grounding constraints:} LLMs classify based on course sheet text, not actual course practice. A course may have excellent GenAI integration in practice but poor documentation in the sheet, yielding a misclassified Clear risk rating.

 \item \textbf{Error inevitability:} At scale, some inconsistency is operationally unavoidable. Based on the 1\% quality sampling of production outputs (\Cref{sec:pipeline}), we estimate true hallucinations \citep{ji2023hallucination}---reports containing claims not present in source documents---at approximately 1\% or lower. However, for a corpus of nearly 5,000 sheets, this still represents a non-trivial number of cases requiring manual review. Additionally, we observed that the same course sheet occasionally received different classifications in separate processing runs, suggesting inherent stochasticity in LLM outputs that institutions should account for when deploying such systems.

 \item \textbf{Ambiguity persistence:} Some cases are genuinely ambiguous; the appropriate classification depends on pedagogical values not captured in the course sheet. LLMs cannot resolve such ambiguity; they can only make it explicit.

 \item \textbf{Language and cultural specificity:} The method's transferability depends on language-specific prompt tuning and cultural understanding of educational contexts (e.g., the Dutch \textit{stage} system differs from Anglophone internships).
\end{enumerate}

\subsection{Transferability to Other Domains}

While our application concerned GenAI risk, the method generalises to other audit domains:

\begin{itemize}
 \item \textbf{Sustainability integration:} Classifying courses by their attention to sustainability learning outcomes;
 \item \textbf{Accessibility compliance:} Assessing whether course descriptions address accessibility for students with disabilities;
 \item \textbf{Assessment diversity:} Auditing over-reliance on particular assessment formats within programmes;
 \item \textbf{Work-integrated learning:} Mapping internship and placement opportunities across the curriculum.
\end{itemize}

Transfer requires: (a) domain-appropriate taxonomy; (b) pilot sample with expert labels; (c) iteration to address domain-specific ambiguities; and (d) validation of error rates against intended use.

\subsection{Key Findings}

Reflecting on the methodological challenges outlined in \Cref{sec:rq}, four themes emerge from our experience.

\textbf{Validation under constraints.} Our five-iteration protocol with a fixed pilot sample and systematic error categorisation provided sufficient validation for institutional deployment. The key insight was prioritising operational validity (fitness for guiding action) over benchmark perfection: agreement thresholds and stopping criteria can be calibrated to the intended use rather than abstract accuracy standards.

\textbf{Error patterns at scale.} Internships and physical/practical components emerged as the primary challenge areas. The model initially over-relied on location (off-campus = high risk) and under-weighted supervision structures. These patterns were not fully captured in the pilot sample, underscoring the difficulty of assembling representative test sets from large, heterogeneous corpora.

\textbf{Longitudinal comparison.} Explicit documentation of prompt changes (in our case, the prompt was held constant between scans), sensitivity analysis with matched subsets, and qualitative review of category changes enable bounded comparison across measurement waves.

\textbf{Responsible deployment.} The combination of pilot iteration, production sampling, uncertainty communication, and stakeholder engagement through direct email provides a workable model for responsible deployment. Critical elements include caveats in all automated communications, clear escalation paths, and separation between guidance and binding policy.

\subsection{Responsible Deployment}

We offer the following practical recommendations for institutions considering similar deployments:

\begin{tcolorbox}[title=Practical Recommendations, colback=green!5!white, colframe=green!75!black]
\textbf{Validation protocol:}
\begin{enumerate}
 \item Assemble a diverse pilot sample ($n \geq 30$) with expert manual labels.
 \item Plan for 4--6 iteration cycles with systematic error categorisation.
 \item Document all prompt changes and their rationale.
 \item Establish stopping criteria (e.g., agreement $\geq 85\%$ or plateau across iterations).
\end{enumerate}

\textbf{Minimal quality-assurance checks:}
\begin{enumerate}
 \item Conduct random sampling (${\sim}$1\%) for manual review before distribution.
 \item Perform automated checks for output format compliance.
 \item Detect gross errors (e.g., hallucinated course components).
 \item Track email bounces during distribution.
\end{enumerate}

\textbf{Uncertainty communication:}
\begin{enumerate}
 \item Include explicit caveats in all automated communications.
 \item Provide clear escalation paths for contested classifications.
 \item Distinguish between automated guidance and official policy.
\end{enumerate}

\textbf{Longitudinal comparison:}
\begin{enumerate}
 \item Maintain version-control prompts and document changes.
 \item Use sensitivity analysis to bound pipeline confounds.
 \item Review category changes qualitatively before quantitative aggregation.
 \item Consider maintaining a fixed ``validation subset'' across waves.
\end{enumerate}
\end{tcolorbox}

\section{Conclusion}

This paper has presented and validated a reusable institutional method for large-scale classification of course information sheets using large language models. The four-phase approach---pilot sampling, iterative prompt engineering, production scanning with stakeholder communication, and longitudinal comparison---enables rapid transformation of heterogeneous HTML catalogue data into structured intelligence for educational governance.

Key contributions include: (1) a documented protocol for prompt iteration under institutional constraints; (2) empirical characterisation of error patterns at scale, particularly around internships and practical requirements; and (3) explicit treatment of pipeline confounds in longitudinal LLM-based analysis. The demonstrated method is transferable to audit domains beyond GenAI risk, offering a template for responsible institutional deployment of generative AI in higher education administration.

The work also illuminates inherent tensions in automated educational analysis: between scale and accuracy, between standardisation and contextual sensitivity, and between efficiency gains and the irreducible need for human judgment. Navigating these tensions requires not merely technical implementation but organisational commitment to using automated intelligence as input to human decision-making rather than its replacement.

\clearpage

\section*{Acknowledgements}
This research was supported by funding from the Flemish Government under the ``Onderzoeksprogramma Artifici\"ele Intelligentie (AI) Vlaanderen'' program.
Andres Algaba acknowledges support from the Francqui Foundation (Belgium) through a Francqui Start-Up Grant and a fellowship from the Research Foundation Flanders (FWO) under Grant No.1286924N. 
Vincent Ginis acknowledges support from Research Foundation Flanders under Grant No.G032822N and G0K9322N. 

\clearpage
\printbibliography

\clearpage

\appendix
\makeatletter
\@addtoreset{equation}{section}
\makeatother
\renewcommand{\theequation}{\arabic{equation}}
\renewcommand{\theHequation}{app.\thesection.\arabic{equation}}

\section{System Prompt (Final Version)}
\label{app:prompt}

\begin{verbatim}
Role: You are an evaluator analyzing university course fiches for potential 
issues related to generative AI usage in coursework and assessments. 
Your objective is to ensure that the course design adequately addresses risks 
associated with generative AI and to offer specific, creative, context-specific 
recommendations for improvement.

The final report is sent on behalf of the AI4education-team and addressed 
directly to the teaching staff as 'jullie' (or the appropriate translation 
if the fiche is in another language).

All responses must be in the language indicated by the fiche's 'onderwijstaal' 
or 'taught in.' When necessary, translate the assigned Impact of GenAI 
("Low risk", "Potential risk", "Clear risk") into that language. 
Sometimes a course has multiple course fiches underneath each other, your 
feedback should always entail all of them!

You always start your report with the overall score, let me clarify:

Overall Score:
  Provide one overall rating with the following clear guidelines:
  - Low risk: No significant risks regarding LLM use.
  - Potential risk: Potential risks identified; requires reflection.
  - Clear risk: Clear risks identified; action needed.
  Your scoring should be nuanced; minor issues may warrant a "Low risk" rating. 
  Note that if nothing is mentioned about the use of AI and there is a risk 
  that students may use it in an uncontrolled setting, the rating should be 
  "Clear risk".

Generative AI Considerations:
  - Identify every section where generative AI is mentioned or should be 
    mentioned. The use of GenAI must be included under the section 'additional 
    information regarding assessment' (this is 'aanvullende informatie mbt 
    evaluatie' in Dutch). If nothing is mentioned here or elsewhere in the 
    course description, recommend making explicit whether the use of GenAI is 
    permitted or not for any particular assignment.
  - For any course component and assignments taken in a non-controlled setting--
    especially writing assignments or take-home work--where clear guidelines on 
    generative AI usage are absent, consider this a clear risk, especially when 
    this counts for a rather large part of the final mark.
  - If guidelines on AI usage are mentioned in the fiche, restate them and 
    remark that this is positive and, in your initial evaluation, ask whether 
    these guidelines are up to date and effectively available to students via 
    Canvas.

Monitoring & Verification:
  - Prohibited AI Use:
    - If generative AI is prohibited, verify whether the evaluation, assignment 
      or task is taken in a controlled, supervised-setting (= in class, on 
      campus) to exclude the unauthorized use of GenAI.  
      - If no such monitoring is mentioned and these tasks can be made at home 
        without supervision, mark as "Clear risk".
      - If monitoring is vague or unspecific, mark as "Potential risk" and 
        recommend that the evaluation should take place in a controlled setting 
        or that the way of evaluation should be reconsidered.
  - Allowed AI Use:
    - If the use of GenAI is allowed, ask whether the guidelines are explicitly 
      provided on the platform Canvas. If Canvas is mentioned here, mark as 
      "Low risk" (this is good, you may add one or two questions if you doubt 
      the feasibility); if detailed guidelines are available, you should not add 
      further questions.
  - For off-campus or take-home assignments, ask if there is an integrated 
    'AI-check' (e.g., process checkpoints, oral defenses, draft verifications, 
    or reflective logs) explicitly tied to the specific assignments. 
  - For written or practical exams, you can assume that they take place on 
    campus, note that the risk of AI misuse is low here since the exam format 
    inherently prevents secret use of AI. Courses where those parts constitute 
    a large percentage of the final mark should not be classified as "Clear risk".
  - If it is explicitly mentioned in the course description that the use of AI 
    is only limitedly permitted or permitted for specific components, ask if 
    specific instructions are communicated via the learning platform (Canvas) 
    or in the assignment instruction. 
  - When discussing assignments or tasks, always name the specific tasks 
    (e.g., 'Assignment X', 'Seminar Project Y') and reference any percentages 
    or explicit descriptions provided in the fiche. Avoid generalities; be 
    specific about which tasks and what their contribution is to the final grade.

Feedback Style:
  - Write a continuous text for the overall reasoning, in maximum 2 paragraphs.
  - Use bullet points only for listing specific, concrete recommendations or 
    alternative fixes.
  - Good feedback should:
    - Describe the current situation with clear, course-specific references. 
      Specify exactly which assignments or components are at risk, citing their 
      names, percentages, or explicit descriptions provided in the fiche.
    - Identify concrete weaknesses in addressing generative AI risks, avoiding 
      general or vague language.
    - Include reflective questions that directly relate to the provided course 
      information and the achievement of the learning goals. For example, ask 
      whether the assessment methods effectively ensure that students reach the 
      intended learning outcomes without unauthorized AI aid.
    - Suggest context-dependent recommendations based on core ideas such as 
      ensuring transparency and verifiability of the work process and verifying 
      individual student effort. Where relevant, suggest concrete, fitting 
      changes--specifying exactly which part of an assignment could be modified 
      or supplemented.
  - Do not mention the use of plagiarism detectors or external detection 
    software; AI monitoring must be integrated into the assessment design. 
    Do not mention "control" because there is no external control, either AI 
    use is monitored or it is integrated into the course and its learning 
    outcomes (leerresultaten).  
  - Avoid repeating the same points; be creative and offer ad hoc solutions 
    based on the specific course information.
  - Do not include generic references to external pages; focus solely on the 
    course context and specifics provided.
  - All content in your report must be in the same language as indicated by 
    the fiche. Do not mix languages or include any headings or terms in a 
    different language.
  - Focus solely on aspects related to AI and its impact on the assessment; 
    do not comment on the overall substance of the course.

Tone & Addressing:
  - Address the teaching staff directly as 'jullie' (or the appropriate 
    translation).
  - Avoid excessive or generic praise. Even for a "Low risk" rating, keep 
    feedback modest yet constructive.
  - For "Potential risk" or "Clear risk" ratings, include reflective questions 
    (reflectieve vragen) that help evaluate whether the course's main learning 
    goals are being met and whether AI misuse might unduly aid students in 
    reaching these goals. For example, ask if the assessment methods 
    sufficiently ensure that students achieve the intended learning outcomes 
    without prohibited AI support.
  - Your tone should remain balanced and fair, ensuring that minor issues do 
    not result in overly harsh criticism.

Report Template:

"Impact of GenAI":
State the overall rating ("Low risk", "Potential risk", or "Clear risk"), 
translated into the fiche's language if necessary.

Reasoning:
- For "Low risk":
  Provide 3-4 lines of concise, course-specific feedback confirming that 
  generative AI usage is appropriately addressed or that the impact of GenAI 
  is low for this course design. For example, if a written or practical exam 
  is held on campus with no take-home assignments, note that the risk of AI 
  misuse is low. If guidelines are mentioned, remark positively and, as an 
  initial note, ask whether these guidelines are effectively available to 
  students via Canvas.

- For "Potential risk":
  Provide 2-10 lines of detailed, specific feedback that highlight moderate 
  risks and concerns. Clearly reference which assignments or components 
  (by name, percentage, or description) are affected and explain why the 
  current design might allow potential AI misuse. Add reflective questions.
  Then, offer concrete recommendations that fit the course context.

- For "Clear risk":
  Provide 2-15 lines of detailed, specific feedback that clearly identify 
  significant risks. Clearly indicate which course components are at high risk 
  by referencing their specific names or percentages as provided, and explain 
  why the lack of explicit AI guidelines or integrated monitoring measures is 
  problematic. Add reflective questions. Then, provide concrete, tailored 
  recommendations.

When using Dutch you should provide good translations like:
"Kunnen de huidige AI-richtlijnen en, indien van toepassing, monitoringsmechanismen 
voldoende garanderen dat de studenten de leerdoelen bereiken zonder 
ongeoorloofde hulp van AI?"
or 
"Bevat dit opleidingsonderdeel leerdoelen die met generatieve AI kunnen worden 
behaald, zonder dat de student voldoende eigen inbreng levert?"

Other questions that might be informative:
- Hoeveel van het eindcijfer wordt bepaald door opdrachten zonder toezicht?
- Stellen de opdrachten de docent in staat de voortgang van de student naar 
  het eindproduct te volgen? 
- Stellen de opdrachten de studenten in staat om te slagen zonder vakspecifieke 
  kennis of vaardigheden?

You should be a bit flexible and not just repeat the examples above. 

Language Requirement:
Always respond in the language indicated by the fiche's 'onderwijstaal' or 
'taught in.' When using a language other than English, sculpt your translation 
with care, always provide neat text that is well-translated.
\end{verbatim}

\section{Example Course Information Sheet}
\label{app:example-sheet}

The following is an excerpt from a representative course information sheet (translated from Dutch), showing the typical structure and content processed by the classification system. Full sheets are publicly available at \url{https://caliweb.vub.be/}.

\begin{tcolorbox}[title=Example Course Information Sheet (Excerpt), colback=white, colframe=black!50]
\textbf{Course Title:} Psychologie en Behavioural Medicine\\
\textbf{Credits:} 4 ECTS\\
\textbf{Study Time:} 108 hours\\
\textbf{Semester:} 2nd semester\\
\textbf{Language of Instruction:} Dutch\\
\textbf{Faculty:} Faculty of Medicine and Pharmacy\\

\textbf{Teaching Team:}
\begin{itemize}[noitemsep]
    \item Cleo Crunelle
    \item Liesbeth Santermans  
    \item Yasmina Bellen
    \item Nathalie Vanderbruggen (coordinator)
    \item Dieter Zeeuws
    \item Michiel Van Kernebeek
    \item Nina Vanden Driessche
\end{itemize}

\textbf{Assessment:}
\begin{itemize}[noitemsep]
    \item Written examination (100\% of final grade)
    \item Second examination opportunity: Yes
    \item Evaluation method: Assessment (0 to 20)
\end{itemize}

\textbf{Course Description:}
This course provides an introduction to psychology and behavioural medicine, 
integrating psychological principles with medical practice. Students learn 
to assess and address psychological factors in health and illness.
\end{tcolorbox}

\section{Example Generated Report}
\label{app:example-report}

The following is a representative example of a generated report (course name anonymised). This example was classified as ``Potential risk'' and demonstrates the structure and tone of the feedback provided to teaching teams.

\begin{tcolorbox}[title=Example Generated Report, colback=white, colframe=black!50]
\textbf{``Impact of GenAI'': Potential risk}

\textbf{Reasoning:}
The course includes a clear mention of generative AI usage, requiring students 
to adhere to the general guidelines set by VUB. Students must explicitly state 
and reflect on their use of generative AI in their assignments. This is a 
positive step towards transparency. However, the course relies heavily on 
process and product evaluations, including project reports and final papers, 
which are completed outside of a controlled environment. This presents a 
potential risk for unauthorized AI use, as these assignments significantly 
contribute to the final grade. While the guidelines are mentioned, it is 
crucial to ensure they are effectively communicated and accessible to students 
via Canvas.

\textbf{Recommendations:}
\begin{itemize}[noitemsep]
    \item Ensure that the guidelines on generative AI usage are prominently 
          available on Canvas and are up-to-date.
    \item Consider integrating checkpoints such as draft submissions or oral 
          defenses to monitor the progress of assignments and ensure authenticity.
    \item Reflect on whether the current assessment methods sufficiently 
          safeguard the learning outcomes from being achieved with unauthorized 
          AI support.
    \item Clarify if there are specific components where AI use is limited or 
          permitted, and communicate these instructions clearly to students.
    \item Explore the possibility of incorporating reflective logs or annotated 
          drafts to verify individual contributions and understanding.
\end{itemize}
\end{tcolorbox}

\end{document}